%% file: iclr2023_conference.tex
\documentclass{article} 
\usepackage{iclr2023_conference,times}

\input{math_commands.tex}

\usepackage{hyperref}
\usepackage{url}
\usepackage{graphicx,subfig}
\usepackage{booktabs}

\title{ANAct: Adaptive Normalization for Activation Functions}

\iclrfinalcopy

\author{Yuan Peiwen \\
School of Computer and Communication\\
Lanzhou University of Technology\\
Lanzhou, China \\
\texttt{yuanpw@lut.edu.cn} \And 
Henan Liu \\
CRRC Zhuzhou Instiute\\
Zhuzhou, China \\
\texttt{liuhenan@gmail.com} \And 
Zhu Changsheng \thanks{Corresponding author. Email: \texttt{zhucs@lut.edu.cn}} \\
School of Computer and Communication\\
Lanzhou University of Technology\\
Lanzhou, China \\
\texttt{zhucs@gmail.com} \And 
Yuyi Wang \thanks{Corresponding author. Email: \texttt{wangyuyi920@gmail.com}} \\
CRRC Zhuzhou Instiute\\
Zhuzhou, China \\
\texttt{wangyuyi920@gmail.com}
}

\begin{document}

\maketitle

\begin{abstract}
	In this paper, we investigate the negative effect of activation functions on forward and backward propagation and how to counteract this effect. First, We examine how activation functions affect the forward and backward propagation of neural networks and derive a general form for gradient variance that extends the previous work in this area. We try to use mini-batch statistics to dynamically update the normalization factor to ensure the normalization property throughout the training process, rather than only accounting for the state of the neural network after weight initialization. Second, we propose ANAct, a method that normalizes activation functions to maintain consistent gradient variance across layers and demonstrate its effectiveness through experiments. We observe that the convergence rate is roughly related to the normalization property. We compare ANAct with several common activation functions on CNNs and residual networks and show that ANAct consistently improves their performance. For instance, normalized Swish achieves 1.4\% higher top-1 accuracy than vanilla Swish on ResNet50 with the Tiny ImageNet dataset and more than 1.2\% higher with CIFAR-100.
\end{abstract}

\section{Introduction}
Deep neural networks \citep{Alex12, He2016, DBLP:conf/nips/VaswaniSPUJGKP17, DBLP:conf/naacl/DevlinCLT19} have attained great empirical success across computer vision, natural language processing, and speech tasks. It should be partly attributed to the decades of research to understand the difficulty of training a deep neural network and proposed solutions.

Various initialization \citep{Glorot2010, DBLP:journals/corr/SaxeMG13, He2015, DBLP:journals/corr/KrahenbuhlDDD15} methods are proposed to help with the convergence of deep models. Xavier initialization \citep{Glorot2010} was purposed to keep the variance of the gradient for all weight matrices in order to mitigate the problem of vanishing and exploding gradients. Since Xavier initialization is designed for symmetric activation function, its derivation only considers the activation function with unit derivative in an idealized initial state. \citet{He2015} has taken account of ReLU in an idealized initial state and gives the specified initialization strategy for neural networks using ReLU and ReLU-like activation functions. However, we would prefer to have a unified method instead of deriving different initializations for different activation functions. More importantly, these works solely consider the initial state of a network and, therefore, the effectiveness may shrink rapidly during training. 

Additionally, GPN (Gaussian-Poincaré normalized) function \citep{DBLP:journals/corr/abs-2006-12169} is a related work proposed to normalize the norm of the output and the derivative of the activation function with the goal of preventing vanishing and exploding gradients. The purpose of their approach is different from ours, to keep the norm of the forward output and the backward pseudo-output same as the norm of the forward input and backward pseudo-input, which is more similar to Xavier initialization. And our approach is applicable in residual networks and can work without the constraint on the variance of the input. 

In this work, we analyze the impact of activation functions on gradients and introduce a theoretically sound approach for normalizing activation functions. A motivation for our approach comes from dropout \citep{JMLR:v15:srivastava14a} loosely. During the back-propagation, ReLU performs like dropout. Nonetheless, in contrast to ReLU, the outputs of dropout are scaled by a factor to recover the mean, which inspired us. Our contributions can be listed as:
\begin{itemize}
	\item We introduce a unified method to normalize different activation functions without derivation for different initialization.
	\item Our method works well relatively regardless of the change from the initial state due to the normalization factor which is updated dynamically during training.
	\item We investigate its compatibility with BN (Batch Normalization) \citep{DBLP:conf/icml/IoffeS15} and residual networks \citep{He2016} and find normalized ReLU and normalized Swish can improve the performances for kinds of networks.
\end{itemize}
\section{Approach}
\index{definitions}
First, We analyze the impact of activation functions on convergence. Then, we demonstrate our approach to normalize activation functions derived from the former part. 
\subsection{the Impact of Activation Functions on Gradients}
\label{subsection:actfunc-impact}
Consider a $N$-layer network with weight matrices $\mW_n$, bias vectors $\vb_n$, activation function $\delta_n$, preactivations $h_n$ and postactivations $x_n$. Assume $x_0$ is the input of the network and $d_n$ is the input size of layer $n$. We can say that,
\begin{align}
	\label{eq:activation}
	&\ x_n = \delta_n(h_n) \\
	\label{eq:linearity}
	&\ h_n = \mW_n^T x_{n - 1} + \vb_n 
\end{align}
where the $x_{n - 1}$ is the input of $n$-th layer and $x_n$ is the output of $n$-th layer.

According to these definitions, we can obtain the equation below in a linear regime, which is similar to the formula in \citet{Glorot2010}:
\begin{align}
	\label{eq:wGrad}
	&\ \Var\left[\frac{\partial Cost}{\partial \mW_n}\right] = 
	\left(\prod_{i = 1}^{n - 1}{d_{i} \Var\left[\mW_i\right]} \right)
	\left(\prod_{i = n + 1}^N{d_{i + 1} \Var\left[\mW_i\right]} \right) \times \Var\left[x_0\right] \Var\left[\frac{\partial Cost}{\partial x_N}\right] 
\end{align}
Then following Xavier initialization we use to constrain weight matrices, we can derive that:
\begin{align}
	\label{eq:xavierInit}
	&\ \forall n, \Var\left[\mW_n\right] = \frac{2}{d_n + d_{n + 1}}
\end{align}
Now, let us take the activation function into consideration. Define $\rho_i$, $\rho_i'$ as:
\begin{align}
	\label{eq:epsilon_i}
	&\ \rho_n = \frac{\E_{h_n \sim \mathcal{N}(0,  \sigma^2)}\left[\delta_n(h_n)^2\right]}{\Var_{h_n \sim \mathcal{N}(0, \sigma^2)}[h_n]} \\
	\label{eq:actfuncGrad}
	&\ \rho_n' = \E_{h_n \sim \mathcal{N}(0,  \sigma^2)}\left[\left(\frac{d \delta_n(h_n)}{d h_n} \right)^2\right]
\end{align}
Note that we assume all $h_n$ are approximately zero-mean Gaussian. By combining them into \ref{eq:wGrad}, we have
\begin{align}
	\begin{split}
		\label{eq:wGradWithActFunc}
		\Var\left[\frac{\partial Cost}{\partial \mW_n}\right] = 
		& \left(\prod_{i = 1}^{n - 1}{\rho_i d_{i} \Var\left[\mW_i\right]} \right)
		\rho_n' \left(\prod_{i = n + 1}^N{\rho_i' d_{i + 1} \Var\left[\mW_i\right]} \right) \\
		& \times \Var\left[x_0\right] \Var\left[\frac{\partial Cost}{\partial x_N}\right]
	\end{split}
\end{align}
For the reason that we use Xavier initialization that normalizes the weight matrices, we can loosely simplify \ref{eq:wGradWithActFunc} into:
\begin{align}
	\begin{split}
		\label{eq:simpleWGradWithActFunc}
		\Var\left[\frac{\partial Cost}{\partial \mW_n}\right] = 
		& \prod_{i = 1}^{n - 1}{\rho_i} 
		\prod_{i = n}^N{\rho_i'}
		\times \Var\left[x_0\right] \Var\left[\frac{\partial Cost}{\partial x_N}\right]
	\end{split}
\end{align}
In order to make the variance of the gradient on each layer approximately same to achieve better convergence, we would like the activation function to satisfy an interesting property as feasible:
\begin{align}
	\label{eq:bidirectionRatio}
	\forall i, \rho_i \approx \rho_i' \approx 1
\end{align}
In fact, this indicates two properties implicitly: (i) $\rho_i \approx 1, \rho_i' \approx 1$ (ii) $\rho_i \approx \rho_i'$. Property (ii) is inherent to an activation function, whereas we can make it more satisfy property (i) by normalizing $\rho_i$ and $\rho_i'$.
\subsection{Approach}
In order to normalize $\rho_i$ and $\rho_i'$, let us apply a normalization factor $\lambda_i$ to the post-activation of the activation function $\delta_i$. From equation \ref{eq:simpleWGradWithActFunc}, we would have:
\begin{align}
	\begin{split}
		\label{eq:simpleWGradWithActFuncWithNormalizationFactor}
		\Var\left[\frac{\partial Cost}{\partial \mW_n}\right] = 
		& \prod_{i = 1}^{n - 1}{\lambda^2_i\rho_i} 
		\prod_{i = n}^N{\lambda^2_i\rho_i'}
		\times \Var\left[x_0\right] \Var\left[\frac{\partial Cost}{\partial x_N}\right]
	\end{split}
\end{align}
From the perspective of forward-propagation, it is expected that:
\begin{align}
	\label{eq:bidirectionRatioFP}
	\forall i, \lambda^2_i\rho_i = 1
\end{align}
From the perspective of backward-propagation, we would expect to have:
\begin{align}
	\label{eq:bidirectionRatioBP}
	\forall i, \lambda^2_i\rho_i' = 1
\end{align}
Since $\rho_i$ and $\rho_i'$ given an activation function $\delta_i$ can be calculated based on its input of the current batch in the period of forward-propagation, we take the reciprocal of their harmonic mean for $\lambda^2_i$ as a compromise, of which the strategy is similar to yet slightly different from Xavier initialization, between preceding two constraints:
\begin{align}
	\label{eq:normalizationFactor}
	\forall i, \lambda_i\ = \sqrt{\frac{\rho_i + \rho_i'}{2 \rho_i \rho_i'}}
\end{align}
As we know, the output of each layer can easily keep zero-mean in a linear regime. However, asymmetric activation functions distort the distribution of output from zero mean and the normalization factor further deteriorates the distortion. Additionally, the equation \ref{eq:wGradWithActFunc} which underlies all of our derivations rests on a fundamental assumption that the weight matrices are zero-mean, that can not be ensured during training due to \emph{internal covariate shift} (ICS) \citep{DBLP:conf/icml/IoffeS15}. In order to inhibit the distortion, we shift $x_{i - 1}$, the post-activation of $\delta_{i - 1}$, to zero-mean in order to obtain zero-mean gradient on the weight matrix. And at the same time, the post-activation fixed to zero-mean further stabilizes the condition of equation \ref{eq:wGradWithActFunc} for
\begin{align}
	\label{eq:EwGrad}
	&\ \mathbb{E}\left[\frac{\partial Cost}{\partial \mW_i}\right] = 
	\mathbb{E}\left[x_{x - i}\right]\mathbb{E}\left[\frac{\partial Cost}{\partial h_i}\right] 
\end{align}
Of course, it requires the assumption that $x_{i - 1}$ is independent of $\frac{\partial Cost}{\partial h_i}$.
\setlength{\belowcaptionskip}{0pt}
\begin{figure}
	\centering
	\subfloat[Vanilla Activation Function]{\includegraphics[width=.40\linewidth]{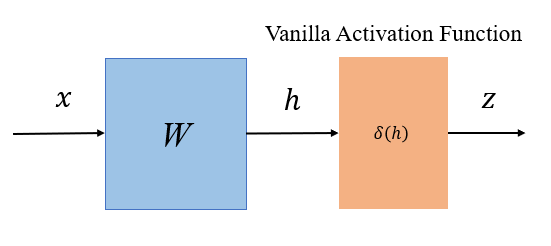}}\vspace{2pt}
	\subfloat[Normalized Activation Function]{\includegraphics[width=.70\linewidth]{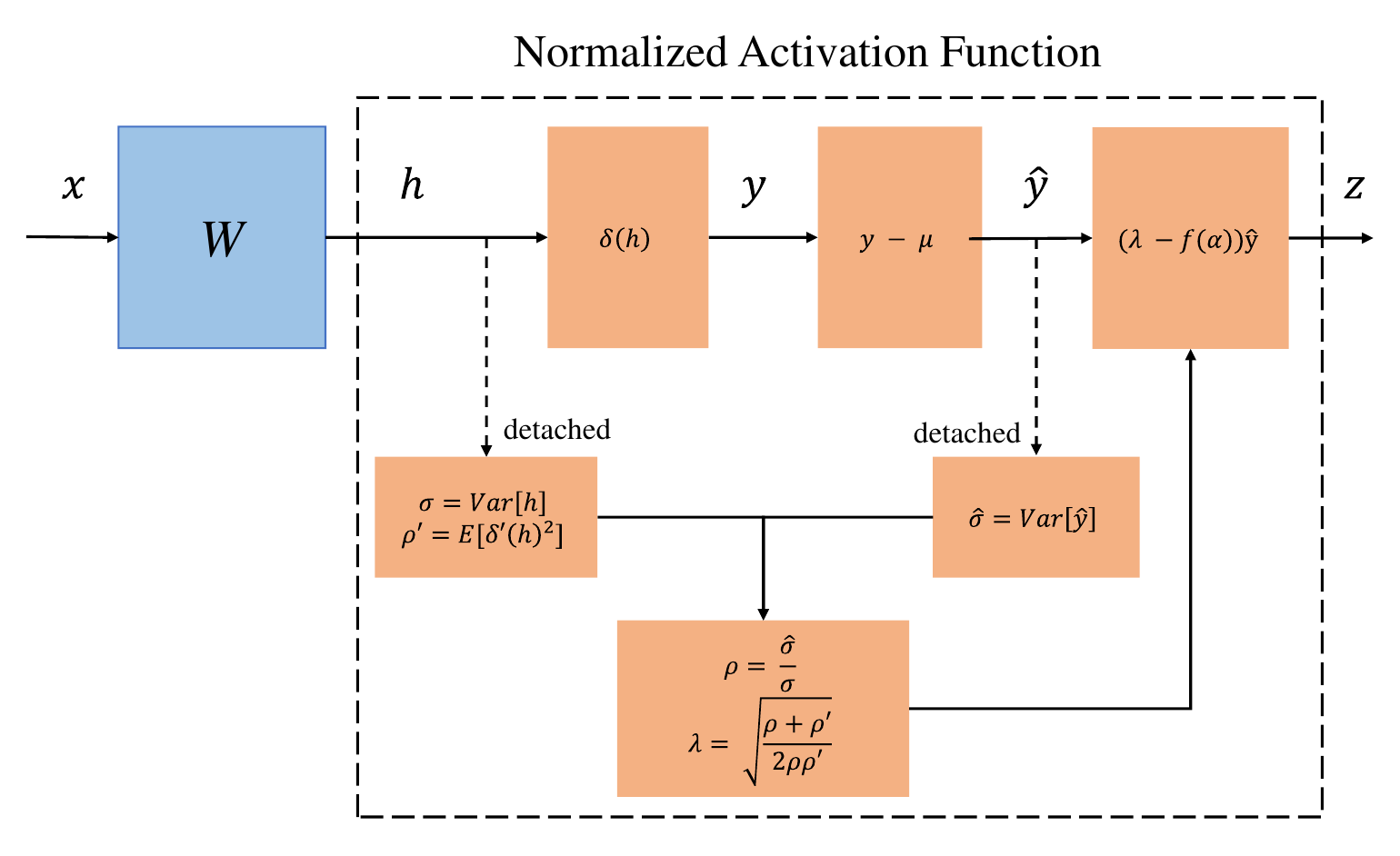}}
	\caption[Vanilla Activation Function And Normalized Activation Function]{The upper panel and the lower panel are two different activation function architectures we compare: (a) Vanilla Activation Function (b) Normalized Activation Function}
	\label{fig:comparison}
\end{figure}

At present, we can give the formula of our approach to normalize $\rho$ and $\rho'$ for given activation function $\delta$:
\begin{align}
	\label{eq:bidirectionRatioFunction}
	\hat{\delta}\left(x\right) = 
	& \left(\lambda + f\left(\alpha\right)\right) \left(\delta\left(x\right) - \mu\right) 
\end{align}
where $\mu$ is the expectation of $\delta\left(x\right)$; $\lambda$ denotes the normalization factor; $f$ is a bounded function to adjust $\lambda$; $\alpha$ is a learnable parameter. We call $\hat{\delta}\left(x\right)$ the \textbf{normalized activation function}. In this paper, we use $f\left(\alpha\right) = 0.3 \mathrm{Tanh}\left(\alpha\right)$ and $\alpha = 0$ as initialization in all experiments. 

In order to control the noise, $\rho$, $\rho'$ and $\mu$ are updated by a momentum parameter $m$ based on history and current mini-batch. And they are also filtered out abnormal values out of bounds with two hyperparameters $L$ and $U$. The updating calculation can be described as the following:
\begin{align}
	\label{eq:muUpdate}
	\mu^{(t)} = 
	& m \mu_M + (1 - m) \mu^{(t - 1)} \\
	\label{eq:rhoUpdate}
	\rho^{(t)} = 
	& \left\{
	\begin{aligned}
		\rho_M, &  \mbox{if } t = 0 \\
		m \rho_M + (1 - m) \rho^{(t - 1)}, &  \mbox{if } L\rho^{(t - 1)} < \rho_M < U\rho^{(t - 1)} \\
		\rho^{t - 1}, &  \mbox{otherwise } \\
	\end{aligned}
	\right.\\
	\label{eq:rho2Update}
	\rho'^{(t)} = 
	& \left\{
	\begin{aligned}
		\rho'_M, &  \mbox{if } t = 0 \\
		m \rho'_M + (1 - m) \rho'^{(t - 1)}, &  \mbox{if } L\rho'^{(t - 1)} < \rho'_M < U\rho'^{(t - 1)} \\
		\rho'^{(t - 1)}, &  \mbox{otherwise } \\
	\end{aligned}
	\right.
\end{align}
where $t$ denotes the number of batches (or iterations).

Note that $\rho$, $\rho'$ and $\mu$ are obtained without gradient calculation. It means they are treated as three constants during backward-propagation. Therefore, this setting bypasses the problem of blocking the first and second derivatives of the loss that batch normalization operation suffers from \citep{Zhou2022BatchNI}.

As a result that $x$ can be roughly guaranteed zero-mean as the assumption, we evaluated several prevalent activation functions by analytically calculating $\mathcal{R}$ score defined below.
\begin{align}
	\label{eq:rScore}
	\mathcal{R}_{x \sim \mathcal{N}
		\left(0, \sigma\right)}\left(\delta, \sigma\right) 
	&=  \ln \frac{\mathbb{E}
		\left[\delta(x)^2\right] - \mathbb{E}\left[\delta(x)\right]^2
	}{\Var
		\left[x\right]
		\mathbb{E}
		\left[\left(\frac{\partial \delta\left(x\right)}{\partial x} \right)^2\right]}  \\
	&\ =  \ln \frac{ 
		\int_{-\infty}^{\infty}{\delta^2(x) \omega(x) \mathrm{d}x} - \frac{1}{\sigma\sqrt{2 \pi}}\left(\int_{-\infty}^{\infty}{\delta(x) \omega(x) \mathrm{d}x}\right)^2
	}
	{
		\sigma^2 
		\int_{-\infty}^{\infty}{\left(\frac{\partial \delta(x)}{\partial x}\right)^2 \omega(x) \mathrm{d}x}	
	} \\
	\label{eq:omega}
	\omega\left(x\right) &= \exp\left(-\frac{x^2}{2 \sigma^2}\right)
\end{align}
\setlength{\belowcaptionskip}{0pt}
\begin{figure}
	\centering
	\includegraphics[width=0.5\columnwidth]{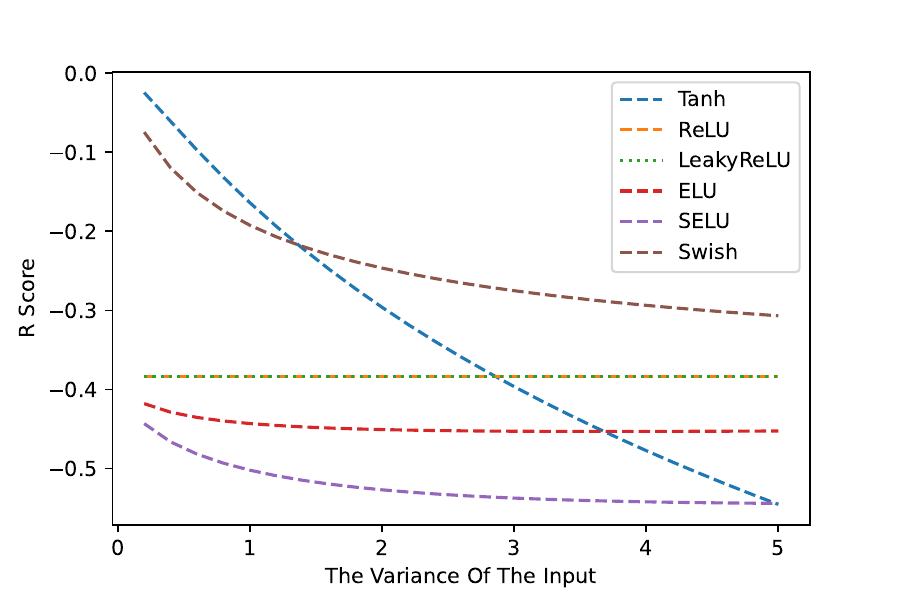}
	\caption[R score]{As shown above, the $\mathcal{R}$ score of common activation functions are constantly below 0. The $\mathcal{R}$ score of Tanh and Swish is desirably near 0 when the variance of input is small. However, they suffer from pseudo-linearity \citep{DBLP:conf/iclr/PhilippSC18} at the same time. The $\mathcal{R}$ score of ReLU and LeakyReLU are fixed at -0.383. Data are calculated with Mathematica.}
	\label{fig:rScore}
\end{figure}
In a normalized activation function, $\rho$ and $\rho'$ can be normalized to be approximately around 1. Namely, property (i) is approached. Nonetheless, prevalent activation functions fail to achieve property (ii) as shown in figure \ref{fig:rScore}. You may ask whether we can find a nonlinear activation function of which $\mathcal{R}$ converges to 0 and the post-activation is zero-mean. The answer is no. \citet{DBLP:journals/corr/abs-2006-12169} has proven it when the input is standard Gaussian. We further generalize their conclusion to any zero-mean Gaussian distribution and detailedly prove the following proposition in Appendix.

Proposition: Assume $x \sim \mathcal{N}\left(0, \sigma\right)$ and function $\delta: \R \rightarrow \R$ , then $\E\left[\delta\left(x\right)\right] = 0$ and $\frac{\E\left[\delta\left(x\right)^2\right]}{\sigma^2} = \E\left[\delta'\left(x\right)^2\right] = C$ if and only if $\delta\left(x\right) = \pm \sqrt{C}x$.

\section{Experiments}
\label{experiments}
With several experiments, we validate the effectiveness and advantage of our approach and investigate its compatibility with BN and residual networks. We benchmark NReLU (normalized ReLU) and NSwish (normalized Swish) against common activation functions, especially their unnormalized counterparts, as baselines. For convenience, we call a convolutional/linear layer plus the following BN and activation function a \emph{super-layer} as a whole in this section.

All activation functions we compare are listed in Appendix. 
\subsection{LeNet5}
First, we compare NReLU and NSwish against all baseline activation functions on LeNet5 using MNIST as the dataset. We run experiments in 50 epochs 25 times for each activation function and use the same learning rate with SGD. Networks using Tanh, NReLU and NSwish are initialized with Xavier initialization; ReLU, LReLU, ELU \citep{DBLP:journals/corr/ClevertUH15}, SELU \citep{DBLP:conf/nips/KlambauerUMH17} and Swish \citep{hendrycks2016gaussian, DBLP:conf/iclr/RamachandranZL18} with He initialization; ReLU-GPN \citep{DBLP:journals/corr/abs-2006-12169} with orthogonal initialization \citep{DBLP:journals/corr/SaxeMG13}. We compare the mean and median of the accuracy and the number of models that fail to reach 98\% validation accuracy until the 5-th, 10-th, 15-th, 30-th and 50-th epoch. Results are shown in Table \ref{tab:mnist}.
\begin{table}[ht]
	\caption[Comparison of activation functions on MNIST dataset]{The table shows the best validation accuracy of different activation functions and the number of models that fail to reach 98\% validation accuracy until different epochs. The best method among unnormalized/GPN/normalized versions is marked with ``*''.
		\label{tab:mnist}}
	\centering
	\begin{tabular}{lccccccc}
		\toprule
		$\ \ $ &
		\multicolumn{2}{c}{validation accuracy} &
		\multicolumn{5}{c}{n models under 98\% accuracy until {\em i}-th epoch}\\
		method & mean & median & 5-th$\ \ $ & 10-th$\ \ $ & 15-th$\ \ $ & 30-th & 50-th \\ 
		\midrule 
		Tanh        & 98.83$\ \ $ & 98.83$\ \ $ & 25 & 9 & 0 & 0 & 0 \\ 
		LReLU & 98.87$\ \ $ & 98.87$\ \ $ & 25 & 4 & 0 & 0 & 0 \\ 
		ELU  & 98.90$\ \ $ & 98.90$\ \ $ & 23 & 0 & 0 & 0 & 0 \\ 
		SELU    & 98.90$\ \ $ & 98.91$\ \ $ & 9 & 0 & 0 & 0 & 0 \\ 
		\cmidrule(r){2-8}
		ReLU  & 94.42$\ \ $ & 98.92$\ \ $ & 19 & 15 & 14 & 11 & 11 \\ 
		ReLU-GPN  & 87.50$\ \ $ & 98.91$\ \ $ & 14 & 7 & 6 & 5 & 5 \\
		NReLU  & {\bf 98.95}* & {\bf 98.96}* & 0 & 0 & 0 & 0 & 0 \\
		\cmidrule(r){2-8}
		Swish    & 98.86$\ \ $ & 98.85$\ \ $ & 25 & 16 & 0 & 0 & 0 \\
		NSwish    & 98.90* & 98.90* & 1 & 0 & 0 & 0 & 0 \\
		\bottomrule
	\end{tabular}
\end{table}
\begin{figure}[ht]
	\begin{center}
		\subfloat{
			\includegraphics[width=0.47\columnwidth]{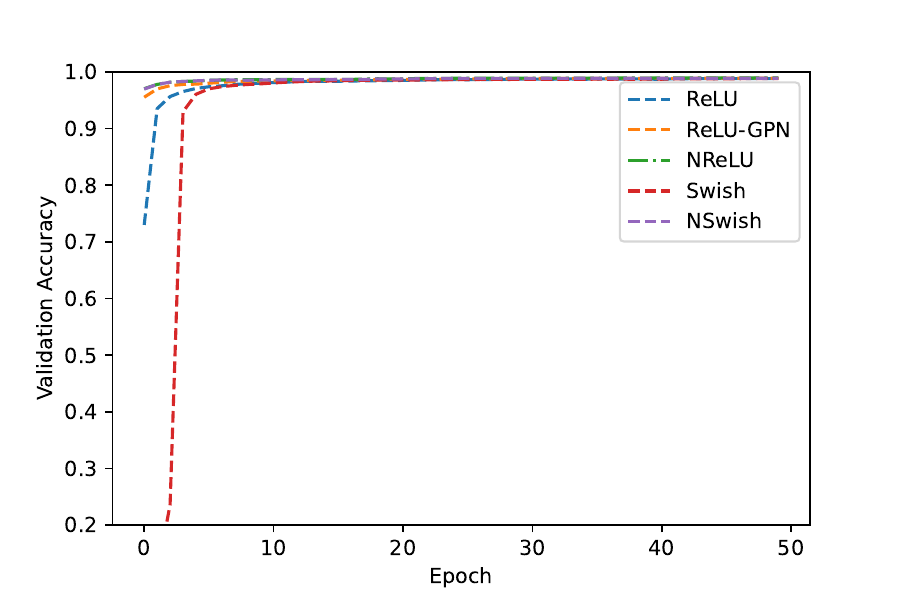}
		}\hfill
		\subfloat{
			\includegraphics[width=0.47\columnwidth]{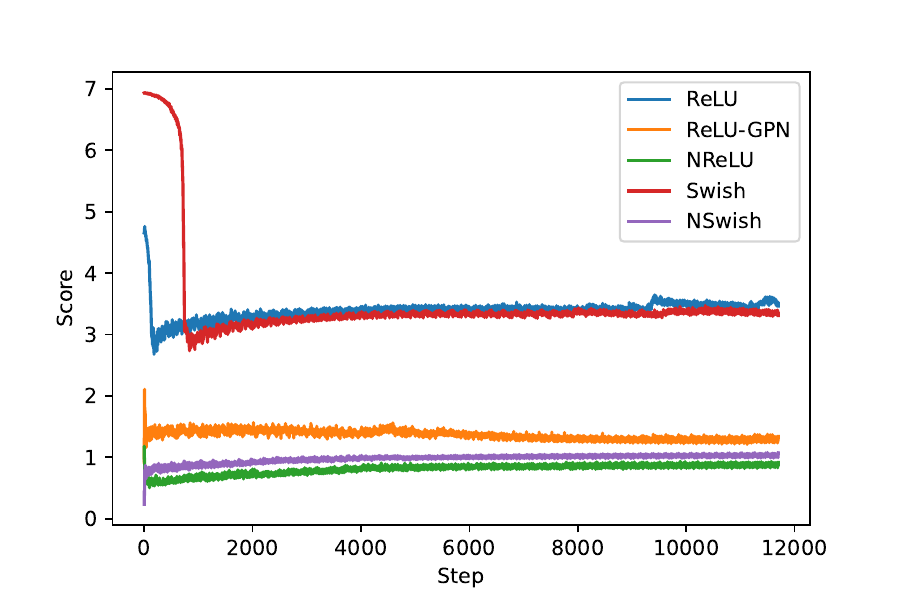}
		}\hfill
	\end{center}
	\caption{The {\bf left} figure shows the increasing validation accuracy.The {\bf right} figure illustrates the scores that we defined in Eq \ref{eq:convergence-score}. Each curve is the median of 25 runs.}
	\label{fig:acc-score}
\end{figure}

Then, we validate how the relation between $\rho$ and $\rho'$ affects the convergence. Namely, how activation functions in a network satisfy property (i) influences the speed of convergence to some extent. 

We recorded the $\rho$ and $\rho'$ during training and compare the score defined below among different methods. 
\begin{align}
	\label{eq:convergence-score}
	&\ Score = \sum_{i = 1}^{N}\frac{\lvert\ln \rho_i\rvert + \lvert\ln \rho_i'\rvert}{2}
\end{align}
By comparison, We find that this score is roughly inversely related to the speed of convergence. We plot the accuracy and scores in Figure \ref{fig:acc-score}. It embodies the impact of activation functions on convergence we mentioned in section \ref{subsection:actfunc-impact}.

\begin{figure}[ht]
	\begin{center}
		\subfloat{
			\includegraphics[width=0.47\columnwidth]{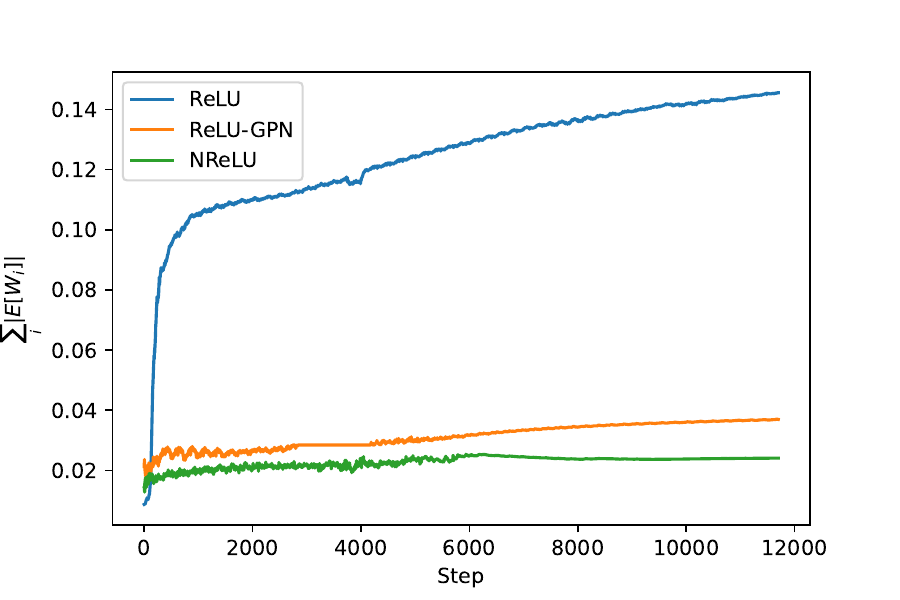}
		}\hfill
		\subfloat{
			\includegraphics[width=0.47\columnwidth]{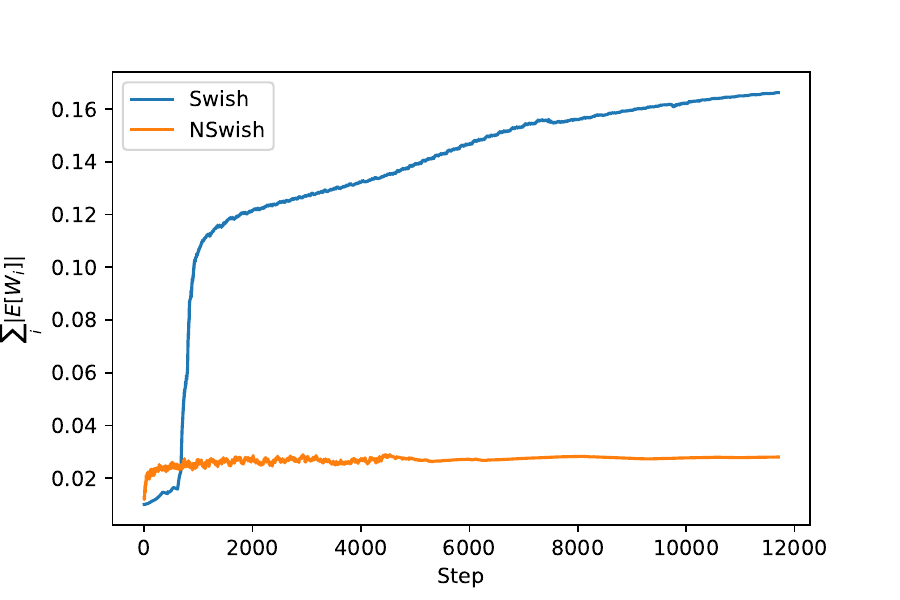}
		}\hfill
	\end{center}
	\caption{We recorded $\sum_i^N \left|\mathbb{E}\mW_i\right|$ during training. The curves are the median of 25 runs.}
	\label{fig:ICS}
\end{figure}

We also monitored the mean of weight matrices. The result in Figure \ref{fig:ICS} shows our approach can inhibit weight matrices to shift from zero mean.

\subsection{VGG}
We benchmark our method on VGG19 \citep{DBLP:journals/corr/SimonyanZ14a} with CIFAR-100 dataset \citep{krizhevsky2009learning}. We replaced ReLU with all activation functions we compare and train for 200 epochs. Super-layers using Tanh, NReLU, NLReLU and NSwish are initialized with Xavier initialization; ReLU, LReLU, ELU, SELU and Swish with He initialization; ReLU-GPN and LReLU-GPN with orthogonal initialization. We follow the same learning rate with AdamW \citep{DBLP:conf/iclr/LoshchilovH19}. For networks using normalized activation functions, we use their unnormalized version as top three activation functions so that we need not change the model architecture and we remove affine transformation in BN followed by a normalized activation function. 

\begin{table}[ht]
	\begin{minipage}[b]{0.49\linewidth}
	\centering
	\small
	\begin{tabular}{lcc}
		\toprule
		$\ \ $ &
		\multicolumn{1}{c}{Top-1 Acc. (\%)} &
		\multicolumn{1}{c}{Top-5 Acc. (\%)}\\
		\midrule 
		Tanh        & 25.63 & 55.50 \\ 
		LReLU & 65.65 & 84.69 \\ 
		ELU  & 59.92 & 85.13  \\ 
		\cmidrule(r){2-3}
		ReLU  & 65.65 & 85.50 \\ 
		NReLU  & 66.11 & 87.13 \\
		\cmidrule(r){2-3}
		Swish    & 64.05 & 85.93 \\
		NSwish    & 65.72 & 87.40 \\
		\bottomrule
	\end{tabular}
\end{minipage}
\hfill%
\begin{minipage}[h]{0.49\linewidth}
	\centering
	\includegraphics[width=\textwidth]{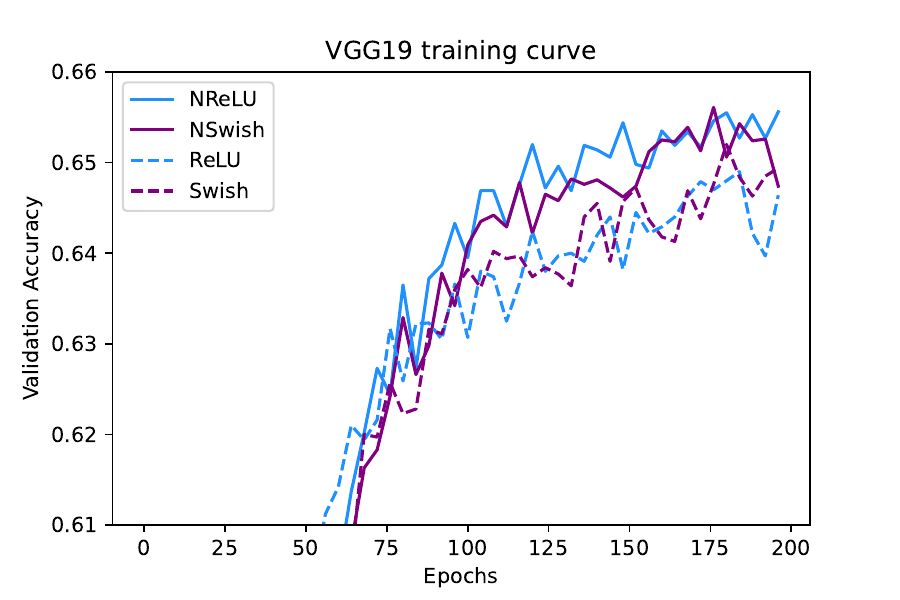}
\end{minipage}%
\\[7pt]
\begin{minipage}[t]{0.49\linewidth}
	\caption{Comparing activation functions on CIFAR-100 dataset using VGG19 as backbone by reporting the median of 3 runs.}
	\label{tab:vgg19-cifar-100}
\end{minipage}%
\hfill%
\begin{minipage}[t]{0.49\linewidth}
	\captionof{figure}{Validation accuracy of VGG19 on CIFAR-100. All curves are the median of 3 runs.}
	\label{fig:vgg19-cifar-100}
\end{minipage}
\end{table}

The results in Table \ref{tab:CIFAR-100} show our approach consistently outperforms its unnormalized counterparts, particularly in terms of top-5 accuracy. NSwish outperforms Swish by a 1.7\% in terms of top1-accuracy.

Our approach makes the gradient on weights less easy to vanish when the network is enough deep so that it improves the trainability of neural networks. The operation of the normalized activation function is similar to BN (batch normalization) operation containing a scale and a shift. Intuitively, we wonder whether these similar operations mutually deteriorate effectiveness. In this experiment, we empirically testify that BN works well with normalized activation functions. 

We find there are two key points when normalized activation functions are used with BN in this experiment.
\begin{itemize}
	\item If we use BN right before normalized activation function, BN without affine transformation will be preferable, whereas removing affine transformation from the super-layer using an unnormalized activation function slightly degenerates the performance. We consider the reason is the affine transformation impairs the effort that BN tries to stabilize the zero-mean assumption and the variance of pre-activation.
	\item It is suggested to use a BN layer between the highest layer using normalized activation function and the top layer as a buffer, typically when normalized activation function is used with BN in lower layers. The reason we consider is that normalized activation function tries to keep the variance of output as the input which is not necessarily same as the variance of the target distribution. This buffer BN can prevent lower layers to raise the variance of output due to the increasing distance among classes during training and thus the moderate change of the variance gives rise to the relatively stable normalization factor.
\end{itemize}
The first point enables us to use fewer parameters, however, to achieve better performance. 

\begin{table}[ht]
	\caption[Comparison of activation functions on CIFAR-100 dataset]{Validation accuracy of VGG19 without BN on CIFAR-100 across the 3 runs ordered by top-1 accuracy.
		\label{tab:CIFAR-100}}
	\centering
	\begin{tabular}{lcccccc}
		\toprule
		$\ \ $ &
		\multicolumn{3}{c}{Top-1 Acc.} &
		\multicolumn{3}{c}{Top-5 Acc.}\\
		\midrule 
		SELU        & 64.49 & 64.16 & 63.83 & 87.28 & 87.31 & 87.07  \\ 
		\cmidrule(r){2-7}
		ReLU-GPN & 64.03 & 63.78 & 63.40 & 86.04 & 86.29 & 86.31  \\ 
		NReLU  & 67.28 & 66.31 & 66.02 & 87.65 & 88.00 & 88.15  \\ 
		\cmidrule(r){2-7}
		LReLU-GPN & 64.09 & 63.73 & 63.28 & 86.03 & 85.87 & 86.06  \\ 
		NLReLU  & 66.33 & 66.22 & 66.06 & 88.09 & 88.42 & 87.89  \\ 
		\cmidrule(r){2-7}
		NSwish  & 64.66 & 63.19 & 63.15 & 85.49 & 85.47 & 86.21 \\
		\bottomrule
	\end{tabular}
\end{table}

For SELU, ReLU-GPN and LReLU-GPN, we run experiments using a learning rate of 0.01 with SGD and decay the learning rate by a factor of 0.2 every 60 epochs since the original optimizer did not converge. Given that SELU, ReLU-GPN and LReLU-GPN are proposed to be used without BN, we remove all BN layers in VGG and compare them with normalized activation functions working without BN. We additionally tested the NLReLU (normalized LeakyReLU) for comparing with LReLU-GPN.

\subsection{ResNet}
We also compared normalized activation function to other methods on ResNet with CIFAR-100 and Tiny ImageNet - a subset of ImageNet \citep{DBLP:journals/ijcv/RussakovskyDSKS15}. Due to the computational limitation, we choose the ResNet50 as the backbone. We follow the same set up and train for a fixed number of epochs, which can ensure the sufficient convergence of baselines.

For normalized activation functions, we replaced all activation functions which is not following the element-wise addition in the bottleneck with their normalized version, since the shift operation will decrease the ratio of positive elements in output and the variance of output will grow exponentially as the information flows to the deep block due to the scale operation, which will lead to extremely unstable normalization factor, if we use normalized activation function right after the element-wise addition. The unnormalized activation function lowers the variance of the output raised by the element-wise addition, whereas the normalized activation function keeps the variance and thus raises it exponentially as residual networks can be unraveled as hierarchical ensembles of relatively shallow networks \citep{Veit}. 

We show the block for normalized activation function and unnormalized version in Appendix.

\begin{table}[ht]
	\begin{minipage}[b]{0.59\linewidth}
		\centering
		\small
	\begin{tabular}{lcccccc}
		\toprule
		$\ \ $ &
		\multicolumn{3}{c}{Top-1 Acc.} &
		\multicolumn{3}{c}{Top-5 Acc.}\\
		\midrule 
		Tanh        & 61.47 & 61.16 & 61.25 & 86.41 & 85.93 & 86.24  \\ 
		LReLU & 71.13 & 70.33 & 70.01 & 91.40 & 90.63 & 91.07  \\ 
		ELU        & 72.10 & 71.96 & 71.82 & 92.03 & 91.89 & 91.75  \\ 
		SELU        & 69.05 & 68.80 & 68.37 & 90.35 & 90.42 & 90.42  \\ 
		\cmidrule(r){2-7}
		ReLU & 70.23 & 70.19 & 69.90 & 90.98 & 91.02 & 90.80  \\ 
		NReLU  & 71.63 & 71.38 & 70.99 & 91.72 & 91.51 & 91.38  \\ 
		\cmidrule(r){2-7}
		Swish  & 71.71 & 71.60 & 71.16 & 91.54 & 91.51 & 91.44 \\
		NSwish  & 73.17 & 72.83 & 72.72 & 92.15 & 92.18 & 92.25 \\
		\bottomrule
	\end{tabular}
\end{minipage}
\hfill%
\begin{minipage}[h]{0.39\linewidth}
	\centering
	\small
	\begin{tabular}{lcc}
		\toprule
		$\ \ $ &
		\multicolumn{1}{c}{Top-1 Acc.} &
		\multicolumn{1}{c}{Top-5 Acc.}\\
		\midrule 
		Tanh      & 40.50 & 66.88 \\ 
		LReLU & 54.90 & 78.20  \\ 
		ELU      & 54.06 & 77.67 \\ 
		SELU        & 48.66 & 74.18 \\ 
		\cmidrule(r){2-3}
		ReLU & 54.46 & 77.77  \\ 
		NReLU & 56.01 & 78.99 \\ 
		\cmidrule(r){2-3}
		Swish  & 54.93 & 78.39 \\
		NSwish  & 56.55 & 79.25 \\
		\bottomrule
	\end{tabular}
\end{minipage}
\\[7pt]
\begin{minipage}[t]{0.59\linewidth}
	\caption{Validation accuracy of ResNet50 on CIFAR-100 across the 3 runs.}
	\label{tab:ResNet50-CIFAR100}
\end{minipage}%
\hfill%
\begin{minipage}[t]{0.39\linewidth}
	\caption{Comparing the median of validation accuracy on Tiny ImangeNet across 3 runs.}
	\label{tab:ResNet50-tiny-imagenet}
\end{minipage}
\end{table}
As the result shown in Table \ref{tab:ResNet50-CIFAR100} and \ref{tab:ResNet50-tiny-imagenet}, NSwish outperforms Swish by nontrivial 1.4\% at least on Tiny ImageNet and 1.2\% on CIFAR-100. And NReLU outperforms ReLU by 1.1\% at least on Tiny ImageNet and 1.0\% on CIFAR-100. At the same time, the total number of trainable parameters of the normalized model is 15K fewer than the unnormalized one.

\subsection{Natural Language Processing}
We benchmark our approach on IMDb dataset \citep{DBLP:conf/acl/MaasDPHNP11} using TextRCNN \citep{DBLP:conf/aaai/LaiXLZ15}. We train each network 5 times with the same set up from scratch, and the median validation accuracy is 90.2\% for NReLU against 89.3\% for ReLU, 90.1\% for NSwish against 89.4\% for Swish. 
 
We also benchmark our method on the domain of machine translation. We train Transformer \citep{ DBLP:conf/nips/VaswaniSPUJGKP17} models initialized with DeepNorm \citep{DBLP:journals/corr/abs-2203-00555} on IWSLT 2015 De-En dataset and evaluate them on test set with standard BLEU metric. 
Higher layers tend to have smaller variance of the output signal of the residual funtion in a network with residual connections \citep{He2016}. Typically, the variance of the residual function output shrinks by the activation function from a relatively lower layer when the linear layer in FFN is broader, since a single layer has more poweful representation. At the same time, the normalized activation function recover the variance of post-activation which is originally supposed to be reduced by the activation function in a high FFN.
Given that, we additionally train models where unnormalized activation functions are kept in the top k FFN of the encoder and decoder. We show the result in Appendix. Normalized activation function will prevent the residual function to have smaller magnitude if it is used in higher layers. However, it more thoroughly exploits the potential representational capacity of lower layers. Note that, in this experiment, $\lambda$ and $\mu$ are obtained with the instance-wise variance and mean.

\section{Discussion}
We have three tips for using normalized activation function with BN and in residual networks,  which we have talked about detailedly in section \ref{experiments}.
\begin{itemize}
	\item If we use BN right before normalized activation function, BN without affine transformation will be preferable.
	\item We suggest using a BN layer between the highest layer using normalized activation function and the top layer as a buffer when normalized activation function is used with BN in lower layers.
	\item When normalized activation function is applied in residual networks, we suggest to not replacing the unnormalized activation functions right after the element-wise addition in each block.
\end{itemize}

\section{Conclusion}
We propose a theoretically sound approach to normalize activation function. Then by carrying on several experiments, we empirically conclude that NReLU and NSwish consistently surpass the accuracy of their unnormalized version and at least one of them can outperform other activation functions, even with fewer parameters if BN is applied. 

\bibliography{iclr2023_conference}
\bibliographystyle{iclr2023_conference}

\newpage
\appendix
\section{Appendix}
\subsection{Activation Functions to Compare}
\begin{itemize}
	\item Tanh:
	\begin{align}
	f(x) = \frac{\exp(x) - \exp(-x)}{\exp(x) + \exp(-x)}
	\end{align}
	\item ReLU:
	\begin{align}
	f(x) &= \left\{
	\begin{aligned}
		x, &  \mbox{if } x > 0 \\
		0, &  \mbox{if } x \leq 0
	\end{aligned}
	\right.
	\end{align}
	\item ReLU-GPN:
	\begin{align}
		f(x) &= \beta \left\{
		\begin{aligned}
			x, &  \mbox{if } x > 0 \\
			0, &  \mbox{if } x \leq 0
		\end{aligned}
		\right.
	\end{align}
	where $\beta \approx 1.4142$.
	\item LeakyReLU:
	\begin{align}
		f(x) &= \left\{
		\begin{aligned}
			x, &  \mbox{if } x > 0 \\
			\alpha x, &  \mbox{if } x \leq 0
		\end{aligned}
		\right.
	\end{align}
	where $\alpha = 0.01$.
	\item LeakyReLU-GPN:
	\begin{align}
		f(x) &= \beta \left\{
		\begin{aligned}
			x, &  \mbox{if } x > 0 \\
			\alpha x, &  \mbox{if } x \leq 0
		\end{aligned}
		\right.
	\end{align}
	where $\alpha = 0.01$; $\beta \approx 1.4141$.
	\item ELU:
	\begin{align}
		f(x) &= \left\{
		\begin{aligned}
			x, &  \mbox{if } x > 0 \\
			\alpha (\exp(x) - 1), &  \mbox{if } x \leq 0
		\end{aligned}
		\right.
	\end{align}
	where $\alpha = 1.0$.
	\item SELU:
	\begin{align}
		f(x) &= \lambda \left\{
		\begin{aligned}
			x, &  \mbox{if } x > 0 \\
			\alpha (\exp(x) - 1), &  \mbox{if } x \leq 0
		\end{aligned}
		\right.
	\end{align}
	where $\alpha \approx 1.6733$; $\beta \approx 1.0507$.
	\item Swish:
	\begin{align}
		f(x) = \frac{x}{1 + \exp(-x)}
	\end{align}
\end{itemize}

Before we introduce normalized activation functions, let us predefine the following in order to avoid duplication:
\begin{align}
	\label{eq:predefine}
	\rho &= \frac{\Var[y]}{\Var[x]}\\
	\lambda &= \sqrt{\frac{\rho + \rho'}{2 \rho \rho'}}
\end{align}
where $x$ denotes the preactivation; $y$ denotes the output of the unnormalized activation function. The definition of $\rho'$ is dependent on the unnormalized activation function and we define them differently in the following parts.
\begin{itemize}
\item NReLU(Normalized ReLU) is defined as 
\begin{align}
	\label{eq:NReLU}
	f(x_i) &= \left(\lambda + \beta \mathrm{Tanh}(\alpha)\right) \left(y_i - \bar{y}\right) \\
	y_i &= \left\{
	\begin{aligned}
		x_i, &  \mbox{if } x > 0 \\
		0, &  \mbox{if } x \leq 0
	\end{aligned}
	\right.\\
	\rho' &= P\left(x > 0\right)
\end{align}
Note that, the elements of the derivative of ReLU can be thought as a random variable following $Bernoulli(p)$ where $p$ is the ratio of positive elements in $x$. We take advantage of this property when talking about it working with the residual connection.

The gradient of $\alpha$ can be easily derived from the chain rule.
\begin{align}
	\label{eq:alphaGradientNReLU}
	\frac{\partial Loss}{\partial \alpha}
	& = \sum_i\beta \left(y_i - \bar{y}\right) \left(1 - \mathrm{Tanh}(\alpha)^2\right) 
\end{align}
where $\sum_i$ runs over all positions of the feature map.
We also need to compute the gradient with respect to the input feature map during training as following
\begin{align}
	\label{eq:xGradientNReLU}
	\frac{\partial Loss}{\partial x_i} &= \left\{
	\begin{aligned}
		\lambda + \beta\mathrm{Tanh}(\alpha), &  \mbox{if } x_i > 0 \\
		0, &  \mbox{if } x_i \leq 0
	\end{aligned}
	\right.
\end{align}
\item NSwish(Normalized Swish) is defined as 
\begin{align}
	\label{eq:NSwish}
	f(x_i) &= \left(\lambda + \beta \mathrm{Tanh}(\alpha)\right) \left(y_i - \bar{y}\right) \\
	y_i &= x_i \mathrm{Sigmoid}(x_i) \\
	\rho' &= \mathbb{E}\left[\left(y + x \mathrm{Sigmoid}(x) \left(1 - y\right)\right)^2\right]
\end{align}
For reason that Swish is a ReLU-like activation function, we also can roughly deem the elements of its derivative as a random variable following $Bernoulli(p)$. The gradient w.r.t. the parameter and the input feature map during training can be derived as:
\begin{align}
	\label{eq:alphaGradientNSwish}
	\frac{\partial Loss}{\partial \alpha}
	& = \sum_i\beta \left(y_i - \bar{y}\right) \left(1 - \mathrm{Tanh}(\alpha)^2\right) \\
	\label{eq:xGradientNSwish}
	\frac{\partial Loss}{\partial x_i} &= \left(\lambda + \beta\mathrm{Tanh}(\alpha)\right) \left(y + x \mathrm{Sigmoid}(x) \left(1 - y\right)\right)
\end{align}
\end{itemize}
\subsection{Proof}
Proposition: Assume $x \sim \mathcal{N}\left(0, \sigma\right)$ and function $\delta: \R \rightarrow \R$ , then $\E\left[\delta\left(x\right)\right] = 0$ and $\frac{\E\left[\delta\left(x\right)^2\right]}{\sigma^2} = \E\left[\delta'\left(x\right)^2\right] = C$ if and only if $\delta\left(x\right) = \pm \sqrt{C}x$.

Proof. Let Hermite polynomials of $k$ degree be:
\begin{align}
	\label{eq:hermitePolynomial}
	H_k(x)
	& = \frac{(-1)^k}{\sqrt{k!}}\exp\left(\frac{x^2}{2\sigma^2}\right)\frac{d^k}{d x^k}\exp\left(-\frac{x^2}{2\sigma^2}\right)
\end{align}
Then we can derive that 
\begin{align}
	\label{eq:orthogonalHermitePolynomial1}
	\int_{-\infty}^{\infty}H_k(x)H_j(x)\exp(-\frac{x^2}{2\sigma^2})
	& = \left\{
	\begin{aligned}
		\sqrt{2 \pi}\sigma^{-(2k - 1)}, &  \mbox{if } k = j \\
		0, &  \mbox{if } k \neq j 
	\end{aligned}
	\right. \\
	\label{eq:orthogonalHermitePolynomial2}
	\int_{-\infty}^{\infty}H'_k(x)H'_j(x)\exp(-\frac{x^2}{2\sigma^2})
	& = \left\{
	\begin{aligned}
		k \sqrt{2 \pi}\sigma^{-(2k + 1)}, &  \mbox{if } k = j \\
		0, &  \mbox{if } k \neq j
	\end{aligned}
	\right.
\end{align}
Since $\E_{x \sim \mathcal{N}\left(0, \sigma\right)}\left[\delta\left(x\right)^2\right] < \infty$; $\E_{\rx \sim \mathcal{N}\left(0, \sigma\right)}\left[\delta'\left(x\right)^2\right] < \infty$ and $\delta\left(x\right)$ and $\delta'\left(x\right)$ can be expanded in terms of Hermite polynomials, we have
\begin{align}
	\label{eq:expansionOfHermitePolynomial}
	\delta(x)
	& = \sum_{k = 0}^{\infty} a_k H_k(x) \\
	\delta'(x)
	& = \sum_{k = 1}^{\infty} a_k H'_k(x)
\end{align}
Due to $\E_{x \sim \mathcal{N}\left(0, \sigma\right)}\left[\delta\left(x\right)\right] = 0$, we have
\begin{align}
	\label{eq:a0zero}
	a_0 = 0
\end{align}
According to Equation \ref{eq:orthogonalHermitePolynomial1}, \ref{eq:orthogonalHermitePolynomial2} and
\begin{align}
\label{eq:condition}
\frac{\E\left[\delta\left(x\right)^2\right]}{\sigma^2} = \E\left[\delta'\left(x\right)^2\right] = C
\end{align}
we have
\begin{align}
	\frac{\E\left[\delta\left(x\right)^2\right]}{\sigma^2} 
	= \frac{1}{\sigma^2} \sum_{k = 1}^{\infty} \sigma^{-2k} a_k^2
	= \E\left[\delta'\left(x\right)^2\right] 
	= \sum_{k = 1}^{\infty} k \sigma^{-2k - 2} a_k^2
	= C
\end{align}
Thus we can derive that
\begin{align}
	\sum_{k = 1}^{\infty} k \sigma^{-2(k + 1)} a_k^2 - 
	\sum_{k = 1}^{\infty} \sigma^{-2(k + 1)} a_k^2
	= 0
\end{align}
that is 
\begin{align}
	\sum_{k = 2}^{\infty} (k - 1) \sigma^{-2(k + 1)} a_k^2
	= 0
\end{align}
For the reason that all terms in $\sum_{k = 2}^{\infty} (k - 1) \sigma^{-2(k + 1)} a_k^2$ is nonnegative, the only solution is $a_k = 0$ for all $k \geq 2$. And for $\E\left[\delta'\left(x\right)^2\right] = \sigma^{-4} a_1^2 = C$, we have $a_1 = \pm \sigma^2 \sqrt{C}$. Hence $\delta(x) = \pm \sqrt{C} x$.

This proof is largely based on \citet{DBLP:journals/corr/abs-2006-12169} with minor generalization here.

\subsection{Block for ResNet50}
\begin{figure}[htb]
	\begin{center}
		\centering
		\subfloat{
			\includegraphics[width=0.40\columnwidth]{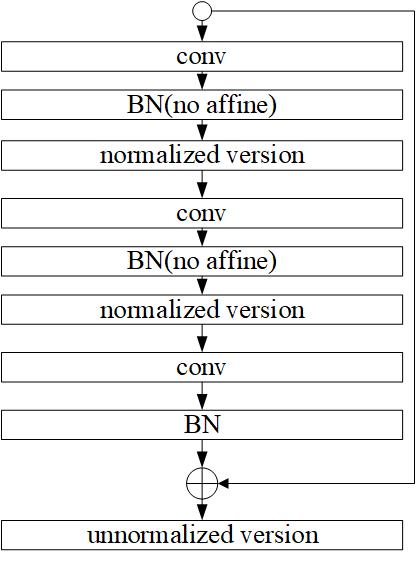}
		}\hfill
		\centering
		\subfloat{
			\includegraphics[width=0.40\columnwidth]{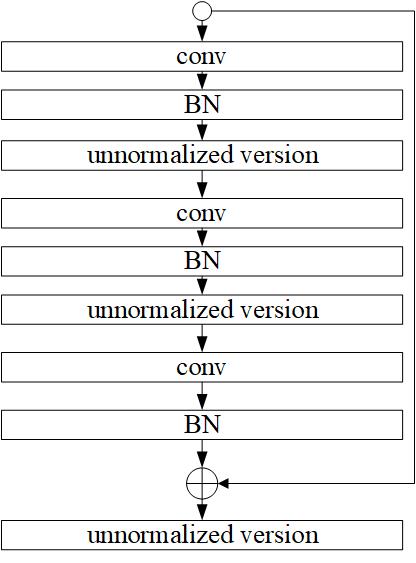}
		}\hfill
	\end{center}
	\caption{{\bf Left}: a block for normalized activation function.{\bf Right}: a block for unnormalized activation function.}
	\label{fig:resnet50-block}
\end{figure}

\subsection{Results And Hyperparamters}

\begin{table}[ht]
	\caption[Comparison of activation functions on IWSLT 2015 dataset]{The BLEU points of 12L-12L Transformer on IWSLT 2015 across different hyperparameter settings. NSwish+$k$Swish denotes that the model uses normalized activation functions as the unnormalized ones are kept in the top $k$ FFN of the encoder and decoder. BLEU scores in parenthesis are calculated with sentences translated with a beam size of 5. The rest experiments are still training at the time of submission.
		\label{tab:IWSLT-2015-Swish}}
	\centering
	\begin{tabular}{lccc}
		\toprule
		$\ \ $ &
		\multicolumn{3}{c}{the size of the first layer in FNN} \\
		$\ \ $ & 512 * 512 & 512 * 1024 & 512 * 2048  \\ 
		\midrule 
		Swish         & 28.38(29.50) & 29.27(30.31) & 29.38(30.52)  \\ 
		NSwish        & 26.78(27.91) & 13.54(14.58) & 11.05(12.98) \\ 
		NSwish+1Swish  & 29.06(30.44) & 29.38(30.17) & 11.43(12.91) \\ 
		NSwish+6Swish  & - & - & 29.37(30.61) \\ 
		\bottomrule
	\end{tabular}
\end{table}

\begin{table}[ht]
	\caption[Comparison of activation functions on IWSLT 2015 dataset]{The BLEU points of 6L-6L Transformer on IWSLT 2015 with different hyperparameter settings. BLEU scores in parenthesis are calculated with sentences translated with a beam size of 5. The $\textit{score with an italic font}$ is the median BLEU of 3 runs. The rest experiments are still training at the time of submission.
		\label{tab:IWSLT-2015-ReLU}}
	\centering
	\begin{tabular}{lccc}
		\toprule
		$\ \ $ &
		\multicolumn{3}{c}{the size of the first layer in FNN} \\
		$\ \ $ & 512 * 512 & 512 * 1024 &  512 * 2048 \\ 
		\midrule 
		ReLU         & \textit{27.69(28.72)} & \textit{27.81(28.92)} & \textit{28.80(29.75)} \\ 
		NReLU+1ReLU  & \textit{27.93(28.81)} & \textit{27.91(29.17)} & - \\ 
		NReLU+3ReLU  & 27.64(28.82) & - &  \textit{29.37(30.43)}  \\ 
		\bottomrule
	\end{tabular}
\end{table}

\begin{table}[ht]
	\caption{The hyperparameter setting for the experiment on VGG19 and ResNet50. Our method works well without warmup as well.
		\label{tab:hyperparameters-vgg-resnet}}
	\centering
	\begin{tabular}{lc}
		\toprule
		Hyperparameters & Value \\
		\midrule 
		Learning Rate & 1e-3 \\
		Batch Size & 128 \\
		Training Epochs & 200 \\
		Warmup Updates  & first epoch \\ 
		Dropout    & 0.5 \\ 
		Gradient Clipping    & 3.0 \\ 
		\bottomrule
	\end{tabular}
\end{table}

\begin{table}[ht]
	\caption{The hyperparameter setting for the experiment on machine translation comparing Swish and Nswish.
		\label{tab:hyperparameters-transformer-swish}}
	\centering
	\begin{tabular}{lc}
		\toprule
		Hyperparameters & Value \\
		\midrule 
		Learning Rate & 5e-4 \\
		Batch Size & 128 \\
		Training Epochs & 20 \\
		Warmup Updates  & 4000 \\ 
		Dropout    & 0.5 \\ 
		Gradient Clipping    & 3.0 \\ 
		Training Max Length  & 50 \\ 
		\bottomrule
	\end{tabular}
\end{table}

\begin{table}[ht]
	\caption{The hyperparameter setting for the experiment on machine translation comparing ReLU and NReLU.
		\label{tab:hyperparameters-transformer-relu}}
	\centering
	\begin{tabular}{lc}
		\toprule
		Hyperparameters & Value \\
		\midrule 
		Learning Rate & 5e-4 \\
		Batch Size & 256 \\
		Training Epochs & 20 \\
		Warmup Updates  & 4000 \\ 
		Dropout    & 0.5 \\ 
		Gradient Clipping    & 3.0 \\ 
		Training Max Length  & 50 \\ 
		\bottomrule
	\end{tabular}
\end{table}

\end{document}

%% file: math_commands.tex
\usepackage{amsmath,amsfonts,bm}

\def\eqref#1{equation~\ref{#1}}

\def\1{\bm{1}}

\def\rx{{\textnormal{x}}}

\def\vb{{\bm{b}}}

\def\mW{{\bm{W}}}

\DeclareMathAlphabet{\mathsfit}{\encodingdefault}{\sfdefault}{m}{sl}
\SetMathAlphabet{\mathsfit}{bold}{\encodingdefault}{\sfdefault}{bx}{n}

\newcommand{\E}{\mathbb{E}}

\newcommand{\R}{\mathbb{R}}

\newcommand{\Var}{\mathrm{Var}}

